# Likelihood Computations Using Value Abstraction

**Nir Friedman**
School of Computer Science & Engineering
Hebrew University
Jerusalem, 91904, ISRAEL
nir@cs.huji.ac.il

**Dan Geiger**
Computer Science Department
Technion
Haifa, Israel
dang@cs.technion.ac.il

**Noam Lotner**
School of Computer Science & Engineering
Hebrew University
Jerusalem, 91904, ISRAEL
noaml@cs.huji.ac.il

## Abstract

In this paper, we use evidence-specific value abstraction for speeding Bayesian networks inference. This is done by grouping variable values and treating the combined values as a single entity. As we show, such abstractions can exploit regularities in conditional probability distributions and also the specific values of observed variables. To formally justify value abstraction, we define the notion of safe value abstraction and devise inference algorithms that use it to reduce the cost of inference. Our procedure is particularly useful for learning complex networks with many hidden variables. In such cases, repeated likelihood computations are required for EM or other parameter optimization techniques. Since these computations are repeated with respect to the same evidence set, our methods can provide significant speedup to the learning procedure. We demonstrate the algorithm on genetic linkage problems where the use of value abstraction sometimes differentiates between a feasible and non-feasible solution.

## 1 Introduction

Inference in probabilistic models plays a significant role in many applications. In this paper we focus on an application in genetics: *linkage analysis*. Linkage analysis is a crucial tool for locating the genes responsible for complex traits (e.g., genetically transmitted diseases or susceptibility to diseases). This analysis uses statistical tools to locate genes and identify the biological function of proteins they encode.

Linkage analysis is based on a clear probabilistic model of genetic events. Mapping of disease genes is done by performing parameter optimization to find the genetic map location that maximizes the likelihood of the evidence (i.e., maximum likelihood estimation). This probabilistic inference is closely related to Bayesian network inference.

Our starting point is the VITESSE algorithm [11], a fairly recent algorithm for linkage analysis that implements some interesting heuristics for speeding up computations. These heuristics achieve impressive speedups that allow to analyze linkage problems that could not be dealt with using the prior state of the art procedures. In the language of Bayesian networks these heuristics can be understood as finding *Value abstractions* (reminiscent of the abstractions studied by [15]). These abstractions are found in an evidence-specific manner to save computations for a specific training example.

In the remainder of this paper we review genetic linkage analysis problems. Then we develop a method to find value abstractions that generalizes the ideas of [11] in a manner that is independent of the inference procedure used. We then extend these ideas in combination with clique-tree inference procedures. Finally, we describe experimental results that examine the effectiveness of these ideas.

## 2 Genetic Linkage Analysis

We now briefly introduce the relevant genetic notions that are needed for the discussion below. We refer the reader to [12] for a comprehensive introduction to linkage analysis.

The human genetic material consists of 22 pairs of *autosomal* chromosomes, and a pair of the *sex* chromosomes. The situation with the later pair is slightly different, and we will restrict the discussion here to the autosomal case, although all the techniques we discuss apply to this case with minor modifications. In each pair of chromosomes, one chromosome is the *paternal* chromosome, inherited from the father, and the other is the *maternal* chromosome, inherited from the mother. We distinguish particular loci in each chromosome pair. Loci that are biologically expressed are called *genes*. At each locus, a chromosome encodes a particular sequence of DNA nucleotides. The variations in these sequences are the source of the variations we see among species members. The possible variants that might appear at a particular locus are called *alleles*. In general, the maternal copy and paternal copy of the same locus can be different.

The aim of linkage analysis to construct *genetic maps* of known loci, and to position newly discovered loci with respect to such maps. Genetic maps describe the relative positions of loci of interest (which can be genes, or ge-netic markers) in terms of their genetic distance. This distance measures the probability of *crossovers* between pairs of loci during *meiosis*, the process of cell division leading to creation of *gemetes* (either sperms or egg cells). During the formation of gemetes, the genetic material undergoes recombination, as shown in a schematic form in Figure 1.

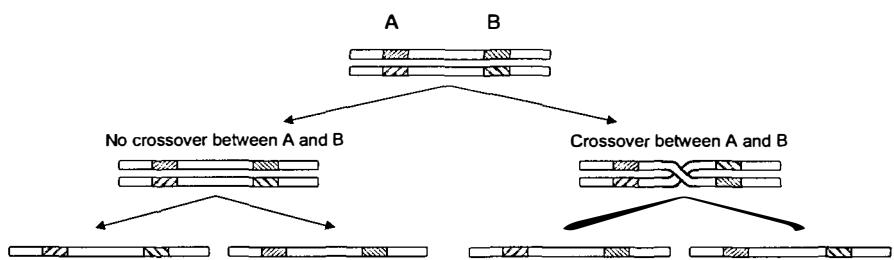

Figure 1: Illustration of recombination during meiosis.

The genetic distance between two loci is measured in terms of the *recombination fraction* between two loci, which is just the probability of recombination of the two loci. The smaller this fraction is, the closer the two loci are. A recombination fraction close to 0.5 indicates that the two loci are sufficiently far so that their inheritance appears independent.

Estimation of these fractions is complicated by the fact that we do not observe alleles on chromosomes. Instead we can observe *phenotype*, which might be traits, such as blood type, eye color, or onset of a disease, or they might oe the result of genetic typing. Genetic typing provides the alleles present in each locus, but does not provide an alignment with the maternal/paternal chromosome. Thus, when genetic typing shows an individual has alleles $a$ and $A$ in one locus, and alleles $b$ and $B$ in another locus, we do not know if $a$ and $b$ where in inherited from the same parent. In such a situation there are 4 four possible configurations ($ab/AB$, $aB/Ab$, $Ab/aB$, $AB/ab$). When we consider multiple loci, the number of possible configurations grows exponentially.

### 2.1 Probabilistic Networks Models of Pedigrees

We start by showing how the underlying model of linkage analysis problems can be represented by probabilistic networks, and then discuss standard approached for computing likelihoods in pedigrees. We note that the representation of pedigrees in terms of graphical models, have been discussed in [6, 8].

A pedigree defines a joint distribution over the *genotype* and *phenotype* of the individuals. We denote the genotype and phenotype of individual $i$ as $G[i]$ and $P[i]$, respectively. The semantics of a pedigree are: given the genotype of $i$'s parents, $G[i]$ is independent from $G[j]$ for any ancestor $j$ of $i$; and given the genotype $G[i]$, the phenotype $P[i]$ is independent of all other variables in the pedigree. We can represent these assumptions on the distribution of $G[i]$ and $P[i]$ by a network where the parents of $G[i]$ are the $G[j]$ and $G[k]$ where $j$ and $k$ are $i$'s parents, and the parent of $P[i]$ is $G[i]$. Not surprisingly, this network has essentially the same topology as the original pedigree.

The local probability models in the network have one of the following forms:

- General population genotype probabilities: $\Pr(G[i])$, when $i$ is a founder.

- *Transmission models*: $\Pr(G[i] \mid G[j], G[k])$ where $j$ and $k$ are $i$'s parents in the pedigree.[1]

- *Penetrance models*: $\Pr(P[i] \mid G[i])$.

This discussion shows that there is a simple transformation from pedigrees to probabilistic networks. This simple transformation obscures many of the details of the pedigree model within the transition and penetrance models. Both of these local probability models are quite complex. We gain more insight into the "structure" of the joint distribution if we model the pedigree at a more detailed level. This can be done in various ways; e.g., [6, 8]. We find it most convenient to use a representation that is motivated by Lander and Green's [9] representation of pedigrees. For this representation we introduce several types of random variables:

**Genetic Loci.** We denote by $A, B, C, \ldots$ the *loci* of interest in the genetic analysis. For example, these can be marker loci and disease loci. For each individual $i$ and locus $A$, we define random variables $A[i,p]$, $A[i,m]$ whose values are the specific value of the locus $A$ in individual $i$'s *parental* and *maternal* haplotypes (chromosomes), respectively. That is, $A[i,p]$ was inherited from $i$'s father, and $A[i,m]$ was inherited from $i$'s mother.

**Phenotypes.** We denote by $F, G, \ldots$ the phenotypes that are involved in the analysis. These might include disease manifestations, genetic typing, or other observed phenotype such as blood types. For each individual $i$ and phenotype $F$, we define a random variable $F[i]$ that denote the value of the phenotype for the individual $i$.

**Selector variables.** Similar to Lander and Green [9], we use auxiliary variables that denote the inheritance pattern in the pedigree. We denote by $S_A[i,p]$ and $S_A[i,m]$ the *selection* made by the meiosis that resulted in $i$'s genetic makeup. Formally, if $j$ and $k$ denote $i$'s father and mother,

---

[1] We make the standard assumption that if individual $i$ is not a founder, then both of her parents are in the pedigree.

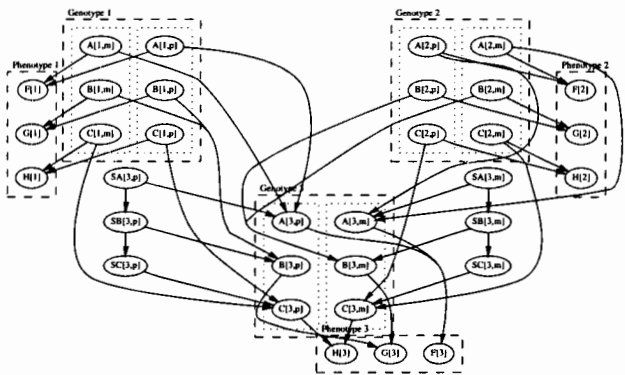

Figure 2: A fragment of a probabilistic network representation of the transmission model, and the penetrance model in a 3-loci analysis.

respectively, then

$$A[i,p] = \begin{cases} A[j,p] & \text{if } S_A[i,p] = 0 \\ A[j,m] & \text{if } S_A[i,p] = 1 \end{cases}$$

and similarly $A[i,m]$ depends on $S_A[i,m]$, $A[k,p]$, and $A[k,m]$.

Using this finer grain representation of the genotype and phenotype we can capture more of the independencies among the variables. For example, $A[i,p]$ and $A[i,m]$ are independent given the genotype of $i$'s parents. Note that, they are dependent given evidence on $i$'s children, or on phenotype that depends on both. Another example, occurs when we know that loci $A$ and $B$ are unlinked (say they are on different chromosomes), then $A[i,p]$ is independent of $B[i,p]$ given the genotype of $i$'s father.

Figure 2 shows a fragment of the network that describes parents-child interaction in a simple 3-loci analysis. The dashed boxes contain all of the variables that describe a single individual's genotypes or phenotype. In this model we assume that loci are mapped in the order $A$, $B$, and $C$. This assumption is reflected in the lack of an edge from $S_A[i,p]$ to $S_C[i,p]$, which implies that the two are independent given the value of $S_B[i,p]$. Figure 2 also shows the penetrance model for this simple 3-loci analysis. In this model we assume that each phenotype variable depends on the genotype at single locus. Again, this is reflected by the fact that each phenotype has edges only from the two haplotypes of a single loci.

### 2.1.1 Likelihood Computation in Pedigrees

There are two main approaches to likelihood computation on pedigrees: Elston-Stewart [3, 5] and Lander-Green [9]. The representation of pedigrees as probabilistic networks allows us to give a unified perspective of both. Broadly speaking, both are variants of variable elimination methods that depend on different strategies for finding elimination ordering, or equivalently, cluster-tree separators.

Figure 3 shows an example of a pedigree. Elston-Stewart's algorithm and later extensions essentially traverse this network along the structure of the family tree.

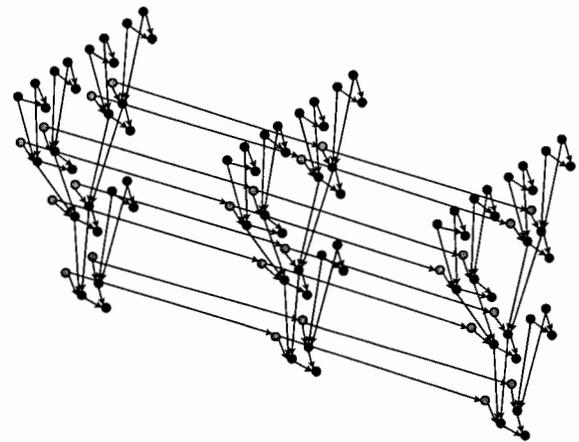

Figure 3: Schematic of a network corresponding to three-loci pedigree. The dark nodes are loci variables in the model (e.g., $A[i,p]$), the dark gray nodes are phenotype variables, and the light gray nodes are selector variables (e.g., $S_A[i,p]$). Each tree-like "slice" corresponds to one locus, and represents the inheritance model for that locus.

At each cluster they aggregate variables that correspond to an individual across all slices. On the other-hand, Lander-Green's algorithm traverses this network from one slice to another. At each step they aggregate all the separator variables at one slice. In this sense, Lander-Green's algorithms treats a pedigree as a *factorial HMM*.

This discussion makes the (known) properties and restrictions of each procedure visible. When the pedigree has loops, the genotypes of individuals are no longer necessarily separators. Thus, one has to resort to approaches for breaking loops. On the other hand, the Lander-Green procedure is not sensitive to loops in the pedigree. However, their procedure cannot applied for pedigree's with many selector variables in each slice and thus, their algorithm is limited to small pedigrees.

### 2.2 Genetic Mapping

The main task for linkage analysis is identifying the map location of disease genes from pedigree data. The standard approach for performing this analysis is to use non-linear optimization procedures that attempt to maximize the likelihood function. Such procedures evaluate the likelihood in several points that are close to each other and estimate the derivative by examining the differences in likelihood between these points. This approach requires several evaluations of the likelihood.

It is important to note that during this optimization, there are many repeated likelihood computations with respect to the same evidence. Moreover, the only parameters that change are the recombination fractions. That is, the only variables whose conditional probability distribution changes are the selector variables.

These repeated computation have been optimized by various approaches. In particular, current linkage analysis soft-ware perform some amount of *genotype* exclusion [13]. These exclusions use several rules of deduction to determine which genotypes are possible for individuals given their phenotype, or the possible genotypes of their direct relatives.

In addition, several researchers made the observation that maintaining the distinction between some of a these values often does not change the probability of the observations. This fact has been exploited in FASTLINK to combine all marker alleles that do not appear in any typed individual in the pedigree [13]. A more powerful use of this idea has been proposed by O'Connell and Weeks [11] and implemented in the VITESSE program, where there is *localized* allele grouping for each individual in the pedigree. The current VITESSE algorithm applies only to loopless pedigree within the framework of bottom-up Elston-Stewart style variable elimination.

## 3  Safe Value Abstractions

In the next sections we develop theory and algorithms that exploit value abstraction in contexts similar to the genetics linkage analysis problems we describe above.

Let $X$ be a variable with a finite domain $Val(X)$ and a probability distribution function $P(X = x)$. An *abstraction of the domain of $X$* is a collection $A$ of subsets of $Val(X)$ that form a non-trivial partition of $Val(X)$: no set in $\mathcal{A}$ is empty, every two sets in $A$ are disjoint, and the union of all sets in $A$ equals $Val(X)$. For every $v \in Val(X)$, let $v^a$ stand for the set in $A$ that contains $v$. We call $v^a$ the *abstract value* corresponding to $v$. Each abstraction defines a *partition function* $\sigma : Val(X) \to C$ which maps a value $v$ to its abstract value $v^a$ via $v^a = \sigma(v)$. An *abstraction of $X$*, denoted by $X^a$, is a variable with a domain $Val(X^a) = A$ which is an abstraction of $Val(X)$, and a probability distribution function $P^a$ given by

$$P^a(X^a = v^a) = \sum_{\{v \in Val(X) \mid v^a = \sigma(v)\}} P(X = v),$$

or in a shorter notation by,

$$P^a(v^a) = \sum_{v \in v^a} P(v).$$

The set of abstractions for $Val(X)$ forms a natural partial order (or more precisely, a *lattice*) as follows. An abstraction $A_1$ is *finer* than abstraction $A_2$ if every set in $A_1$ is a subset of a set in $A_2$, in which case we also say $A_2$ is *coarser* than $A_1$. An abstraction $A_1$ is *strictly finer (coarser)* than abstraction $A_2$ if $A_1$ is finer (coarser) than $A_2$ and $A_1 \neq A_2$. The *maximal abstraction* consists of one set and the *minimal abstraction* consists of singletons.

A *refinement* of two abstractions $A_1$ and $A_2$ is an abstraction $A$ such that $A$ is finer than $A_1$ and finer than $A_2$. A *tight refinement* of two abstractions $A_1$ and $A_2$ is a refinement $A$ such that every other refinement $A'$ of $A_1$ and $A_2$ is finer than $A$. In other words, suppose that $\sigma^1$ and $\sigma^2$ are the two partition functions defined by abstractions $A_1$ and $A_2$, respectively. Then, the partition function of their tight refinement $A$ is given by $\sigma = \sigma^1 \wedge \sigma^2$ defined such that $\sigma(v) = \sigma(v')$ if and only if $\sigma^1(v) = \sigma^1(v')$ and $\sigma^2(v) = \sigma^2(v')$. Intuitively, the refinement of two partition functions defines a partition function that preserves the distinctions made by both partition functions and introduces no new distinctions. When two abstractions are not related through refinement, they are said to be *incomparable*.

*Evidence* with respect to a variable $X$ is an assertion $e$ that the value observed for $X$ is among a subset $O \subset Val(X)$ of the possible values. When $O$ is a singleton, $X$ is said to have been observed.

An abstraction $\sigma$ of $X$ is *safe* with respect to $e$ if $P(e \mid X = x) = P(e \mid X = x')$ for all $x, x'$ such that $\sigma(x) = \sigma(x')$. That is, the distinctions blurred by the abstraction $\sigma$ do not effect the probability of the evidence. We can always find a safe abstraction, since the *trivial* abstraction that consists of singletons is always safe. Moreover, it is clear that there exist a *maximally safe* abstraction which is the coarsest safe abstraction.

As a simple example, consider a game where a player can bet on a dice outcome and wins if the outcome matches his bets. To formalize, suppose we have three variables *Bet* that can take the values *odd* and *even*, *Dice* that can take the values $1, \ldots, 6$, and $Win$ that can be either *yes* or *no*. Suppose also that we observe that the player won, that is *Win* = *yes*. Clearly, the likelihood $P(Win = yes|Dice)$ does not depend on the distinction between all possible outcomes of the dice. Since the player can only bet on even or odd outcome, the abstraction of values $\{1, 3, 5\}, \{2, 4, 6\}$ is a safe one. This abstraction is clearly the maximally safe abstraction of *Dice*. However, there are many other safe abstractions. Note that in this example, if the dice is fair, then $P(Win, Dice = x)$ is the same for all values of $x$ in the same partition. However, the abstraction is still safe even if the dice is not fair. The point is that we can compute $P(Dice \in \{1, 3, 5\})$ without worrying about the rest of the domain (e.g., probability of various bets, etc.).

This simple example suggests that it suffices to consider an abstraction of the dice when we compute the probability of winning in the betting game. This can lead to saving in the number of operations we perform in our calculations. Such computational savings can be much more drastic when one is presented with many interconnected variables as is the case with Bayesian networks. Let $\mathbf{X} = \{X_1, \ldots, X_n\}$ be a set of variables each associated with a finite domain $Val(X_i)$. Also, let $B$, with a directed acyclic graph $G$, stand for a Bayesian network over $\mathbf{X}$ and let $\mathbf{Pa}(X_i)$ be the parents of each $X_i$ in $B$. An *abstraction* $B^a$ of $B$ is a Bayesian network with the same set of vertices and edges as in $B$, and where each variable $X_i$ is replaced with an abstraction $X_i^a$.

We now want to determine the conditional probability distributions in $B^a$. We start by defining the probability of an abstraction given the "un-abstracted" parents:

$$P(X_i^a = x_i^a \mid \mathbf{pa}(X_i) = \mathbf{u}) =$$

**ALGORITHM ValueAbstract(B,e)**

**Input:** A Bayesian network B and evidence $e$
**Output:** A safe abstraction $B^a$ wrt $e$

**Discard:**
For every $X_i$ in $B$, remove from
$Val(X_i)$ all values that are incompatible
with $e$ (e.g., using arc-consistency algorithm)
**Comment:** Nodes in $e$ remain with one value

**Abstract:**
Set $\Sigma_i := \{ Val(X_i)\}$, for $i = 1, \ldots, n$
Iterate over $X_i$ in reverse topological order:
  Suppose that $\mathbf{Pa}(X_i) = \{X_1, \ldots, X_k\}$.
  1. Set $\sigma_i = \bigwedge \Sigma_i$
    **Comment:** This defines $X_i^a$.
  2. Find partition functions $\sigma_1^i, \ldots, \sigma_k^i$
     of $X_1, \ldots, X_k$ such that
     $P(X_i^a \mid x_1, \ldots, x_k) = P(X_i^a \mid x_1', \ldots, x_k')$
     whenever
     $\sigma_1^i(x_1) = \sigma_1^i(x_1',), \ldots, \sigma_k^i(x_k) = \sigma_k^i(x_k')$.
  3. for each $X_j \in \mathbf{Pa}(X_i)$
     $\Sigma_j := \Sigma_j \cup \{\sigma_j^i\}$

**Construct Tables:**
Iterate over $X_i$ in reverse topological order:
  Abstract the table $P(X_i^a \mid \mathbf{pa}(X_i)^a)$
  according $P(X_i \mid \mathbf{pa}(X_i))$.

Figure 4: Computing Value Abstraction

$$\sum_{x_i \in x_i^a} P(X_i = x_i \mid \mathbf{pa}(X_i) = \mathbf{u})$$

where $x_i^a$ is an abstract value of $x_i$. We say that an abstraction of $X_i$'s parents is *cautious* if $P(X_i^a = x_i^a \mid \mathbf{pa}(X_i) = u) = P(X_i^a = x_i^a \mid \mathbf{pa}(X_i) = u')$ for all values $u$ of $\mathbf{Pa}(X_i)$ that are mapped to the same partition. In this case, we define

$$P(X_i^a = x_i^a \mid \mathbf{pa}(X_i)^a = u^a) = \\ P(X_i^a = v^a \mid \mathbf{pa}(X_i) = u) \text{ for } u \in u^a.$$

Note that the notion of an abstract value of a variable is naturally extended to a set of variables via the Cartesian product of the domains of the individual variables.

An abstracted Bayesian network $B^a$ is a *a (safe and cautious) abstraction* of a Bayesian network $B$ over $\mathbf{X}$ wrt evidence $e$ if $P(e|B) = P(e|B^a)$. It is also *maximal* if for any other Bayesian network $B^{a'}$ with this property either $Val(X_i^{a'})$ is finer than $Val(X_i^a)$ or the two sets are incomparable for all variables $X_i$.

## 4 Finding Safe & Cautious Abstractions

We now describe a simple iterative algorithm, *ValueAbstract*, which finds a safe abstraction of a Bayesian network $B$ with respect to evidence $e$. The algorithm consists of three phases:

**Discard** The algorithm starts by examining the variables in the network, and for each $X_i$, discarding all values that are incompatible with the evidence $e$. This is done via any arc-consistency algorithm as described in the CSP literature.

**Abstract** In this phase the algorithm traverses the network from the leafs upwards and computes cautious abstractions for the parents of each abstracted variable. Since a variable $X_i$ can be a parent of several variables, we need to collect the abstractions that are cautious with respect to each of these children. Thus, during this phase, the algorithm maintains a set of abstractions, $\Sigma_i$, that contains the abstractions required for $X_i$ by $X_i$'s children. When the procedure processes $X_i$, it finds the minimal refinement of all these abstractions of $X_i$. We denote by $\bigwedge \Sigma_i$ the tightest refinement of all the abstractions in $\Sigma_i$.

**Construct Tables** In the last phase the algorithm constructs the conditional probabilities in the abstracted network.

The full algorithm is given in Figure 4.

**Theorem 4.1:** *The network $B^a$ returned by* ValueAbstract *is a safe, cautious abstraction of $B$ wrt $e$.*

Ignoring for the moment the cost of the Discard phase, we see that each iteration (either of Abstract, or Construct Tables phase) examines a single family. The cost of such an iteration can be exponential in the number of parents in the family. Thus, the running time of *ValueAbstract* is *linear* in the number of variables in the network, but exponential in the maximal indegree of the network. We stress, however, that for networks in which conditional probabilities are represented by tables, the running time is linear in the size of the network description (since the description of the conditional probability tables are also exponential in the number of parents).

This implies that the running time of this algorithm is not sensitive to the topology of the network, and the complexity of inference with it. VITESSE [11] (see Section 2) is a specialized version of *ValueAbstract* that yields impressive speedup in likelihood calculations in a genetic analysis domain. Thus, this simple algorithm can often make the difference between feasible and infeasible calculations.

## 5 Message-Specific Abstraction

The *ValueAbstract* algorithm has a desirable property: it is independent of the particulars of the inference procedure we use for computing likelihoods. Thus, it can be applied as a preprocessing step before likelihood computation. The simplicity and low complexity of the algorithm make it attractive.

Nonetheless, there are some regularities that are missed by *ValueAbstract*. First, the main processing is strictly bottom-up: the abstraction is constructed from the leaves of the network upward. However, we note that the first phase (Discard using edge-consistency) can propagate implication of evidence to lower nodes.

Second, we considered each variable separately from the others. This limits us to abstractions that are the Cartesian products of the abstractions of variables in the parents set. Thus, rather than holding abstractions per variable, we may wish to hold abstractions for select groups of variables. Suppose, for example, that we have two binary variables $X$ and $Y$ each with values $\{0,1\}$. Suppose the evidence $e$ is such that $X$ and $Y$ must have had the same values but $e$ does not determine which one of their values. A maximal abstraction of $Val(X) \times Val(Y)$ wrt $e$ is the set $\{eq = \{(0,0), (1,1)\}, neq = \{(0,1), (1,0)\}\}$. There exist no maximal abstractions wrt $e$ for $Val(X)$ or for $Val(Y)$ which are strictly coarser than the original sets of values.

Finally, another opportunity for improvement rests on the observation that rather than holding one abstraction per variable, we can hold several abstractions per variable, so that the likelihood computations of different parts of $e$ can be treated more efficiently. Suppose, for example, that we have a Markov chain $X_1 \to X_2 \cdots \to X_n$ and that $X_1$ and $X_n$ are observed. That is, the evidence $e$ is composed of two parts $e_1$ and $e_n$. Then, for any $X_i$, $1 < i < n$, we can think of two natural abstractions, one is a maximal abstraction wrt $e_1$, and the other is a maximal abstraction wrt $e_n$. To compute the posterior $P(X_i|e)$ one would need the tight refinement of both abstractions. However, to pass messages to its neighbors ala Pearl propagation style, we only need to use one of the abstractions, which in general are coarser than their tight refinement, and thus more efficient.

To deal with these issues, we need to develop abstractions that depend on the details of the of the inference procedure we use. We now address these issues within the context of *cluster-tree* (aka *clique tree*) algorithms [7, 10, 14]. We start with a presentation of one variant of tree-based algorithms. The other variants have slightly different details, but our algorithm can be easily adopted to deal with these.

### 5.1 Clique Tree Propagation Algorithm

Assume that $B$ is a fixed network. A *cluster-tree* for $B$ is a tree over $k$ nodes such that:

- Each node $l$ in the tree is annotated with a *cluster* $\mathbf{C}_l \subseteq \{X_1, \ldots, X_n\}$.
- Each variable $X_i$ is *assigned* to one cluster $\mathbf{C}_{l(X_i)}$ such that $X_i \in \mathbf{C}_{l(X_i)}$ and $\mathbf{Pa}(X_i) \subseteq \mathbf{C}_{l(X_i)}$.
- If $X_i \in \mathbf{C}_l$ and $X_i \in \mathbf{C}_m$, then $X_i \in \mathbf{C}_j$ for any node $j$ on the path from $l$ to $m$.

Let $l$ be a node. By definition, if $X_i$ is assigned to $l$, then $X_i$ and $\mathbf{Pa}(X_i)$ are subsets of $\mathbf{C}_l$. Thus, we can define a function on values $\mathbf{c}_l \in Val(\mathbf{C}_l)$

$$g_l(\mathbf{c}_l) = \prod_{i, l(X_i) = l} P(x_i \mid \mathbf{Pa}(X_i))$$

(If $l$ is a node that is not assigned any variable, then we define $g_l(\mathbf{c}_l) = 1$.)

It is easy to see that by simple rearrangement of products the probability $P(x_1, \ldots, x_n \mid B)$ can be rewritten as:

$$P(x_1, \ldots, x_n \mid B) = \prod_l g_l(\mathbf{c}_l).$$

If we have evidence, say on a set of variables $\mathbf{E}$, we can update the functions to reflect that. For example, if $X_1 = a$. We can multiple $g_{l(X_1)}$ by a function $\mathbf{o}_1(X_1)$ such that $\mathbf{o}_1(x_1) = 1$ if $x_1 = a$ and $\mathbf{o}_1(x_1) = 0$ otherwise. If we update the nodes in this manner for all variables in $\mathbf{E}$ according to the specific evidence $\mathbf{e}$ then

$$P(x_1, \ldots, x_n, \mathbf{e} \mid B) = \prod_l g_l(\mathbf{c}_l).$$

To see this, note that if $x_1, \ldots, x_n$ is consistent with $\mathbf{e}$, then its value is not changed by the modification to the $g_l$. On the other hand, if it is not consistent, then one of the $g_l$ is 0, and thus, the probability of the joint assignment is 0.

Let $l$ and $m$ be adjacent nodes in the tree. We define the *separator* $S_{l,m} = \mathbf{C}_l \cap \mathbf{C}_m$. A separator defines a partitions of the clusters $\{C_1, \ldots, C_k\}$ into two sets: the clusters on the $l$-side of the separator, and the clusters on the $m$-side of the separator. We denote these sets $A_l^{l,m}$ and $A_m^{l,m}$. In addition, we define the sets of variables in both groups of clusters $\mathbf{X}_l^{l,m} = \cup A_l^{l,m}$, and $\mathbf{X}_m^{l,m} = \cup A_m^{l,m}$

The key property of separators is that they allow us to factor the computation of probabilities into two separate cases. Using the properties of the tree, it is easy to show that

**Proposition 5.1:**

$$\begin{aligned} P(\mathbf{e} \mid B) &= \sum_{x_1}, \ldots, \sum_{x_n} \prod_l g_l(\mathbf{c}_l) \\ &= \sum_{s_{l,m} \in Val(S_{l,m})} f_m^{l,m}(s_{l,m}, \mathbf{e}) f_l^{l,m}(s_{l,m}, \mathbf{e}) \end{aligned}$$

where

$$f_l^{l,m}(s_{l,m}, \mathbf{e}) = \sum_{\mathbf{x} \in Val(\mathbf{X}_l^{l,m} - S_{l,m})} \prod_{C_n \in A_l^{l,m}} g_n(\mathbf{c}_n)$$

$$f_m^{l,m}(s_{l,m}, \mathbf{e}) = \sum_{\mathbf{x} \in Val(\mathbf{X}_m^{l,m} - S_{l,m})} \prod_{C_n \in A_m^{l,m}} g_n(\mathbf{c}_n).$$

The key property of this factorization is that it is recursive.

**Proposition 5.2:** *Consider a node $l$ whose adjacent nodes are $m_1, \ldots, m_k$. Then*

$$f_l^{l,m_1}(s_{l,m_1}, e) = \sum_{\mathbf{y} \in Val(\mathbf{C}_l - S_{l,m_1})} g_l(\mathbf{c}_l) \prod_{j > 1} f_{m_j}^{l,m_j}(s_{l,m_j}, e).$$

(1)

Thus, to compute the *message* $f_l^{l,m_1}(s_{l,m_1}, e)$ we need to combine the messages from the other clusters adjacent to $l$ with conditional probabilities that are assigned to $l$ and then sum out all of the variables except these on $S_{l,m_1}$.

Using this recursive rule we can compute the *likelihood* $P(e \mid B)$. We choose a separator $S_{l,m}$. According to Proposition 5.1, all we need do is to compute the messages $f_l^{l,m}$ and $f_m^{l,m}$ and then sum over the values of variables in $S_{l,m}$. To compute these two messages, we apply the recursion rule of Proposition 5.2 until we get to the leafs of the tree. It is easy to see that this procedure is closely related to *variable elimination* algorithms [3, 4, 16], except that we eliminated several variables at each step. Moreover, the structure of the clique tree determines the order of elimination.

In addition to likelihood computations, we can also compute the posterior for every cluster: $P(e, \mathbf{c}_l \mid B)$. We do so by combining the messages from all of $l$'s adjacent nodes:

$$P(e, \mathbf{c}_l \mid B) = g_l(\mathbf{c}_l) \prod_j f_{m_j}^{l,m_j}(s_{l,m_j}, e)$$

Cluster tree algorithm compute such a posterior for each cluster. This can be done efficiently by dynamic programming: for each separator we only need to compute two messages. By appropriate use of dynamic programming all of these messages can be computed in two passes over the tree [7, 14].

### 5.2 Clique Tree Abstractions

Suppose we are given a cluster tree and an evidence $e$. Can we abstract the values of cliques and separators? The ideas from the previous section can be applied here in a straightforward fashion. Let $S_{l,m}$ be a separator. An abstraction $\sigma_l^{l,m}$ of $Val(S_{l,m})$ is *safe* for $f_l^{l,m}(s_{l,m}, e)$ if

$$f_l^{l,m}(s_{l,m}, e) = f_l^{l,m}(s'_{l,m}, e)$$

for all $s_{l,m}$ and $s'_{l,m}$ such that $\sigma_l^{l,m}(s_{l,m}) = \sigma_l^{l,m}(s'_{l,m})$.

To construct a safe abstraction for the message $f_l^{l,m}$, we examine the recursive definition given Proposition 5.2. This definition implies that $f_l^{l,m}$ is a function of $g_l$ and $f_{m'}^{l,m'}$ for nodes $m'$ adjacent to $l$. Thus, if we have safe abstractions to all these terms, we only need to preserves values of

$$\sum_{\mathbf{y} \in \mathbf{C}_l - S_{l,m_1}} g_l(\mathbf{c}_l^a) \cdot \prod_{j>1} f_{m_j}^{l,m_j}(s^a_{l,m_j}, e)$$

We construct these abstraction using a dynamic programming procedure that is analogous to the clique-tree propagation algorithm. The difference is that instead of propagating probabilistic messages we are propagating abstractions. We define two operations on abstractions that are the analogs of message multiplication and of marginalization.

We start by combination of abstractions.

**Definition 5.3:** Suppose that $\sigma_X, \sigma_Y$ are abstractions that are safe with respect to $f(\mathbf{X})$ and $g(\mathbf{Y})$, respectively. (The sets $\mathbf{X}$ and $\mathbf{Y}$ can overlap.) Let $\mathbf{Z} = \mathbf{X} \cup \mathbf{Y}$. The combined abstraction $\sigma = \sigma_X \cdot \sigma_Y$ over $Val(\mathbf{Z})$ is such that $\sigma(\mathbf{z}) = \sigma(\mathbf{z}')$ when $\sigma_X(\mathbf{x}) = \sigma_X(\mathbf{x}')$ and $\sigma_Y(\mathbf{y}) = \sigma_Y(\mathbf{y}')$, where $\mathbf{x}$ and $\mathbf{y}$ are the values of $\mathbf{X}$ and $\mathbf{Y}$ specified by $\mathbf{z}$ (and similarly for $\mathbf{x}'$ and $\mathbf{y}'$). ∎

It is easy to check that if $\sigma_X$ and $\sigma_Y$ are safe for $f(\mathbf{X})$ and $g(\mathbf{Y})$, then $\sigma_X \cdot \sigma_Y$ is safe for $f(\mathbf{X})g(\mathbf{Y})$.

The second operation we need to examine is marginalization. Let $f(\mathbf{X}, \mathbf{Y})$ be a factor. We want to find an abstraction that is safe with respect to $g(\mathbf{y}) = \sum_\mathbf{x} f(\mathbf{x}, \mathbf{y})$. To do so, we need to identify values $\mathbf{y}$ for which we are going to add the same values in the same order.

**Definition 5.4 :** Suppose that $\sigma$ is an abstraction of $Val(\mathbf{X}, \mathbf{Y})$ that is safe with respect to $f(\mathbf{X}, \mathbf{Y})$. We define the abstraction $\sigma \downarrow_\mathbf{Y}$ over $Val(\mathbf{Y})$ so that $\sigma \downarrow_\mathbf{Y} (\mathbf{y}) = \sigma \downarrow_\mathbf{Y} (\mathbf{y}')$ if

$$\sigma(\mathbf{x}, \mathbf{y}) = \sigma(\mathbf{x}, \mathbf{y}') \text{ for all } \mathbf{x}$$

∎

Given these two operations, we can define the abstraction algorithm for clique-trees. We start by computing an abstraction $\sigma_l$ of $g_l(\mathbf{c}_l)$ for each clique $l$. This can be done either by combining abstractions for the conditional distributions of variables that are assigned to $l$ or by first constructing $g_l()$ and then finding the coarsest safe abstraction. The first option can introduce unnecessary distinctions, but can be more efficient.

Next, we define the analog of the recursive rule of Proposition 5.2. Consider a node $l$ whose adjacent nodes are $m_1, \ldots, m_k$. Then,

$$\sigma_l^{l,m_1} = \left( \sigma_l \cdot \sigma_{m_2}^{l,m_2} \cdot \ldots \sigma_{m_k}^{l,m_k} \right) \downarrow_{S_{l,m_1}}$$

To construct the abstraction we perform dynamic programming that determines the abstraction for each message in terms of the abstractions for neighboring separators. This dynamic programming is analogous to the propagation of messages in the probabilistic inference algorithm on clique-trees.

Once we compute the abstraction of the messages we can perform inference. The key saving of the abstraction is that in computation we perform multiplication and addition once for every abstract value. Thus, if $S^a$ is the abstracted version of $S$, the saving in computation in construction of the message on $S$ is $|ValS^a|/|ValS|$.

We can show that the resulting algorithm preserves correctness of inferences.

**Theorem 5.5:** *Inference on the abstracted clique tree computes exactly all queries of the form $P(\mathbf{C}_l \mid \mathbf{e})$ for the evidence $\mathbf{e}$ specified at the construction of the abstraction.*

We note that the cost of the construction of the clique-tree abstractions depends on the cost of the basic operations. In the most naive instantiation, we represent abstractions as tables. In this case, the cost of the operations is exactly the same as the cost of probabilistic computation on the clique-tree.

## 6 Abstractions and 0 values

Our algorithm can be easily extended to exploit an additional "structural" feature in conditional probability distri-butions. If at some stage in the algorithm $f_l^{l,m}(\mathbf{s}_{l,m}) = 0$, then multiplications by this value will always result in 0.

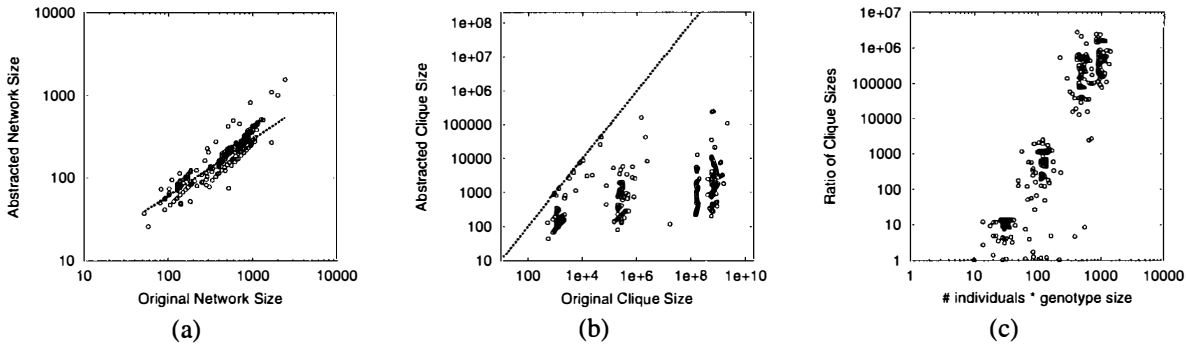

Figure 5: Display of the saving achieved by ValueAbstract on 280 linkage analysis networks. Each point corresponds to a network. Graph (a) shows reduction in network size ($x$-axis is original network size and $y$-axis is reduced network size); Graph (b) shows reduction in clique tree size ($x$-axis is original clique tree size and $y$-axis is the size of the clique tree of the abstracted network); and graph (c) relates the ratio of reduction in clique tree size ($y$-axis) the a complexity estimate of the linkage analysis problem ($x$-axis).

We can record this fact in our abstractions by introducing a special abstract value 0 that corresponds to all the values of the variables that are given a value 0. Then, we modify the definition of combination and marginalization to take the special properties of 0 into account:

- $(\sigma_X \cdot \sigma_Y)(\mathbf{x}) = 0$ if either $\sigma_X(\mathbf{x}) = 0$ or $\sigma_Y(\mathbf{y}) = 0$.
- $\sigma \downarrow_X (\mathbf{y}) = 0$ if $\sigma(\mathbf{x}, \mathbf{y}) = 0$ for all $\mathbf{x}$.

These modifications allow us to deal with evidence more easily. Suppose that a variable $X \in \mathbf{C}_l$ is assigned the value $x$ in the evidence. Then we combine $\sigma_l()$ with an abstraction $\sigma_X$ such that all values $x' \in Val(X) - \{x\}$ are assigned to the abstract value 0, and $x$ is assigned to the singleton set $\{x\}$. When we combine this abstraction with the clique abstraction we ensure that all assignments to $\mathbf{C}_l$ in which $X \neq x$ are assigned to 0.

We note that this simple modification of our procedure essentially implements partial constraint satisfaction propagation to discover unattainable joint assignments to clusters/separators.

## 7 Evaluation

We tested our methods on a collection of standard benchmark pedigrees that are reported in [1, 13]. These pedigrees come from 10 different studies and contain 90 different pedigrees of sizes varying from 5 to 200 individuals. From these we generated 280 different linkage analysis problems by including different numbers of loci in the analysis. These were translated into a Bayesian network of the form described in Section 2.1. For each network we also constructed evidence assignment based on the original findings in the studies and used these in the analysis below.

In the first phase of our experiments we tested the ValueAbstract procedure. This procedure implements the ideas of VITESSE combined with constraint propagation to remove impossible values. Figure 5(a) shows the reduction in the size of the *network* achieved by ValueAbstract. This reduction is due to eliminating and combining values of variables. The reduction in the network size can be approximated as $2.6 \cdot n^{0.68}$ (the line in Figure 5(a)).

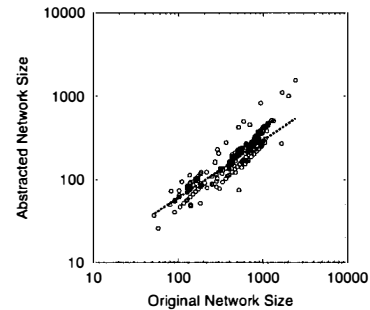

Figure 6: Display of the saving achieved by abstracting values inside the clique tree on the network returned by ValueAbstract. The $x$-axis is the size of the clique tree before abstraction and the $y$-axis is the size of the clique tree after abstraction.

However, since the computation time depends linearly on the size of the clique tree constructed from the network, we also want to measure the reduction in this size. This is shown in Figure 5(b). As we can see, the ratio of reduction can vary significantly. We believe that this is due to structural features of the pedigree. Figure 5(c) shows that the ratio of improvement in the clique tree size is roughly proportional to the product of the number of individuals in the pedigree and the number of genotype values for each individual. This later quantity is a rough estimate of the complexity of the problem.

In the next stage we applied the clique tree abstract procedure described in Section 5. Here we measured the reduction in effective size of the clique tree due to the abstraction of values in cliques and separators. Figure 6 compares the sizes before and after we applied this procedure to the network returned by ValueAbstract. As we can see, we get ad-ditional saving, especially for networks with large cliques. The reduction is estimated as $5.2 \cdot n^{0.64}$ (the line in Figure 6. As we can see, the savings can be drastic.

## 8 Concluding Remarks

In this paper we introduced an approach to exploit regular structure in Bayesian networks to reduce computation time. This approach exploits symmetry to merge values of variables or groups of variables at different stages of the computation. Our motivation is from linkage analysis, where this type of heuristics have been very successful [11]. We are currently extending our implementation to deal with larger networks and plan to incorporate our methods within a linkage analysis software.

It is clear that this approach can be beneficial to other forms of structured Bayesian networks. In particular, networks with CSI [2]. Value abstraction suggests a general framework within which we can evaluate the utility of algorithms that work with tree CPTs. A structured representation of message (i.e., [17]) is essentially an abstraction. If an algorithm is exact, then the representation it uses must be a refinement of the abstraction our algorithm constructs. We plan to exploit this to design "optimal" structure representations for message passing with CSI.

**Acknowledgements**

We thank Tal El-Hay for his help in implementing the Clique tree construction algorithm and Ann Becker and Gal Elidan for help with the benchmark pedigrees. This work was supported by Isreal Science Foundation grant number 224/99-1 and by the generosity of the Michael Sacher fund. Nir Friedman was also supported by Harry & Abe Sherman Senior Lectureship in Computer Science. Experiments reported here were run on equipment funded by an ISF Basic Equipment Grant.